\documentclass{article}

 \usepackage[final]{nips_2018}

\usepackage[utf8]{inputenc} 
\usepackage[T1]{fontenc}    
\usepackage{hyperref}       
\usepackage{url}            
\usepackage{booktabs}       
\usepackage{amsfonts}       
\usepackage{nicefrac}       
\usepackage{microtype}      
\usepackage{graphicx}

\title{Audio Captcha Recognition Using RastaPLP Features by SVM }

\author{
  Ahmet Faruk \c{C}akmak \thanks{https://github.com/cakmakaf/Audio-Captcha-Recognition} \\
  Lecturer Professor \\
  Middle Tennessee State University \\
  Murfreesboro, Tennessee, USA \\
  \texttt{ahmet.cakmak@mtsu.edu} \\
   \And
  Muhammet Balc{\i}lar \\
  Research Engineer\\
  3-Goats\\
  La Courneuve, France \\
  \texttt{msbalci@3-goats.com} \\
}

\begin{document}
\maketitle

\begin{abstract}
Nowadays, CAPTCHAs are computer generated tests that human can pass but current computer systems can not. They have common usage in various web services in order to be able to detect a human from computer programs autonomously. In this way, owners can protect their web services from bots. In addition to visual CAPTCHAs which consist of distorted images, mostly test images, that a user must write some description about that image, there are a significant amount of audio CAPTCHAs as well. Briefly, audio CAPTCHAs are sound files which consist of human sound under heavy noise where the speaker pronounces a bunch of digits consecutively. Generally, in those sound files, there are some periodic and non-periodic noises to get difficult to recognize them with a program but not for a human listener. We gathered numerous randomly collected audio file to train and then test them using our SVM algorithm to be able to extract digits out of each conversation. 
\end{abstract}


\section{Introduction  }

\subsection{Problem Statement}

The term CAPTCHA stands for “Completely Automated Public Turing" test to tell "Computers and Humans Apart” and was invented by von Ahn et al.[\hyperref[sec:r1]{1}]. The current CAPTCHAs can be categorized into three categories: Text-based CAPTCHAs [\hyperref[sec:r2]{2},\hyperref[sec:r3]{3}], Image-based CAPTCHAs [\hyperref[sec:r4]{4},\hyperref[sec:r5]{5}] and Sound-based CAPTCHAs [\hyperref[sec:r6]{6},\hyperref[sec:r7]{7},\hyperref[sec:r8]{8}]. Since CAPTCHAs can individuate between humans and computers with high probability, they are applicable for many various security applications such as preventing bots from voting in online polls, signing in for multiple of spam email accounts, buying tickets to buy an event by itself, etc.  A web user has to properly put down the characters on the screen to be able to pass a common visual CAPTCHA. Although some of the CAPTCHAs have been broken by machine learning techniques [\hyperref[sec:r9]{9},\hyperref[sec:r10]{10}], there are still plenty of them actively secure against attacks. Due to improper visual entries of this type of CAPTCHA, audio CAPTCHAs were invented. A usual audio CAPTCHA consists of single or multiple talkers saying words or digits at some randomly selected time periods. A web client has to properly diagnose the digits or characters pronounced in the audio file to elapse the CAPTCHA. 

In this research, we used RASTA-PLP features and SVM classifier to recognize audio CAPTCHAs. According to our tests, we reached $98\%$ accuracy for individual digit recognition precision, and around $89\%$ accuracy for the entire digit recognition. We provide a classification algorithm to recognize the digits in the data files and obtain the best accuracy via SVM classifiers using Principle Component Analysis (PCA). In regarding of this goal, we compare our results with non PCA Naive Bayes and default SVM classifier.

\subsection{Metrics }

We used two different measurements the first one is the digit success considering the digits are independent of each other, and the second one is the success of complete recognition of all these digits in the test file where they are usually varying between 4 and 6. 

For example, assume that an audio file with a ground truth value of $x=[x_1, x_2, \cdots, x_N]$ is successful on the second measurement if and only if the file recognized it as $\tilde{x}=[\tilde{x}_1, \tilde{x}_2, \cdots, \tilde{x}_N]$, that is, 
\begin{equation}
x_1=\tilde{x}_1, x_2=\tilde{x}_2, \cdots, x_N=\tilde{x}_N.
\end{equation}
Otherwise it will be considered as a failure. However, the digit success measurement is more complicated than the other one. We used Dynamic Time Warping (DTW) algorithm to be able to measure the difference between the order of digit and the order of ground truth digit. In the following sections, we will mention some basic knowledge about DTW and will explain how it helped us to obtain better results. 

DTW [\hyperref[sec:r11]{11}] is an algorithm that calculates an optimal warping path between two time series. The algorithm calculates the distance among two time series and the warping route values between them. The most common used technique for distance is the Euclidian distance between two data elements but we consider the Dirac delta function as our metric. Let's consider the example at the begining of this section again as given by two audio file with a ground truth value of $x=[x_1, x_2, \cdots, x_N]$. That results in a matrix of distances having n lines and m columns of general term so our first performance metric is given by
\[  d_{i} = \left\{
\begin{array}{ll}
      1, & If\,\, m = n\,\, and \,\, x_1=\tilde{x}_1, x_2=\tilde{x}_2, \cdots,x_m=\tilde{x}_n \\
      0, & Othewise \\
\end{array} 
\right. \]
for $i\in \{1,\cdots,N\}$ and the entire prediction accuracy is given by
\begin{equation}
P_{accuracy}=\frac{\sum_{i=1}^{N}d_{i}}{n}
\end{equation}
where $n$ is the total number of test elements.

On the other hand, the minimal distance matrix between sequences is determined using a dynamic programming algorithm and the following optimization process:
\begin{equation}
\ell_{i,j} = D(i,j) +  \min_{i,j} \{ \ell_{i-1,j-1}, \ell_{i-1,j}, \ell_{i,j-1} \}
\end{equation}
where $\ell_{i,j} $ is the minimal distance between the subsequences $x=[x_1, x_2, \cdots, x_n]$ and $\tilde{x}=[\tilde{x}_1, \tilde{x}_2, \cdots, \tilde{x}_n]$.  A warping route is the route along minimal distance matrix from $\ell_{1,1}$ to $\ell_{n,m}$ including those $\ell_{i,j}$ elements that  consist of the $\ell_{n,m}$ distance. The global warp cost of the two sequences can be formalize by
\begin{equation}
Cost = \frac{1}{K}\sum_{i=1}^{K}\omega_{i}
\end{equation}
where $\omega_{i}$ belongs to warping path, and $K$ is the number of them.

The difference function of Dirac delta is 
\[  D(i,j) = \left\{
\begin{array}{ll}
      0, & If\,\, i = j\ \\
      1, & Othewise \\
\end{array} 
\right. \]
Hence, the final calculated value of the cost function of $(Cost)_{i}$ will give us how many digits are NOT equal each other. Therefore the second evaluation metric is the individual digit accuracy must be given by
\begin{equation}
D_{accuracy} = \frac{\sum_{i=1}^{n}(\#\,of\,digits)_{i} - \sum_{i=1}^{n} (Cost)_{i} }{\sum_{i=1}^{n}(\#\,of\,digits)_{i}}.
\end{equation}
That is, $\sum_{i=1}^{n}(\#\,of\,digits)_{i}$ is the total length of digit of the $i^{th}$ element, and $ \sum_{i=1}^{n}(Cost)_{i} $ is the sum of the incompatible digit length.

Roughly speaking, DTW shrinks between two number or string arrays to find the best match. For example, if the number array 123456 was predicted as 2456, then the prediction array would actually overlap with the first array as $*$2$*$456 where $*$ stands for the incorrect predictions of the digits. It is easy to see that since 1 and 3 wasn't predicted correctly, we will say the difference between these two number arrays is 2 and the accuracy of the prediction is $\frac{4}{6}\approx0.66$ or 66\%.

\section{Analysis and Algorithm Techniques}

\subsection{Data Exploration}

We provide features from the audio CAPTCHA  and use SVM classifier to perform automatic speech recognition on segments of the sample files to elapse the audio CAPTCHAs. There exist several well-known methods for extracting specifications from audio files. The technique that we used here is relative spectral transform-perceptual linear prediction (RASTA-PLP). By using RASTA-PLP, we could be able to train our classifiers to identify words and digits all by itself of who pronounce them. 

In [\hyperref[sec:r12]{12},\hyperref[sec:r13]{13}], authors used both PLP and RASTA-PLP could identify particular digits during the existence of noise. However, the noise originating from telephone or microphone tapes in various scenes. In our sample files, we also add vocal or music noise that makes the problem more challenging. In [\hyperref[sec:r10]{10}] Chellapilla et al. struck over how many visual CAPTCHAs can be elapsed by successfully separating the challenge into two minor sub-challenges as segmentation and recognition. Tam et al. [\hyperref[sec:r7]{7}]  analyzed the security of audio CAPTCHAs from popular Web sites by using AdaBoost, SVM, and k-NN, achieved correct solutions for test samples with accuracy up to $71\%$. In addition, the same group at [\hyperref[sec:r14]{14}] analyzed the security of obtaining audio CAPTCHAs and provided a description and analysis of a new and improved audio CAPTCHA.

In this problem, we gathered 900 audio CAPTCHAs from various websites: google.com, digg.com, and recaptcha.net. Each of the CAPTCHAs annotated with the information regarding digit locations provided by the manual transcriptions. For each type of CAPTCHA, we randomly selected 800 samples for training and used the remaining 100 for testing set. The trained samples have been labeled manually one by one. First, we divide the audio file into segments of noise or words to be able to reveal the audio CAPTCHA. In the test files, the words include digits. 

Next, we manage to identify where the locations of the digits in the CAPTCHA file starts and finishes. In 800 trained audio files, the digits 0,1,2,...,8,9 were mentioned loudly 627, 536, 555, 593, 552, 505, 662, 537, 637, 533 times, respectively. Furthermore, all the conversations, where a digit never been pronounced or random noise moments elapsed, have been determined as $11^{th}$ class. We manually obtained 27718 segments of moments which belong to the $11^{th}$ class in the training set.

We can see in Figure 1, a sample audio file is displayed in the time domain and in the frequency domain. In the time domain, the labels of the digits are also shown. In Figure 2, 04648 is pronounced in the sample file. As you can see, many non-numeric regions resemble those in the time domain and the frequency domain. There may be periodic and random noisy background speeches that make automatic identification difficult in non-digitized places.

\begin{figure}\label{Time_Frequency}
  \centering
    \includegraphics[width=0.7\textwidth]{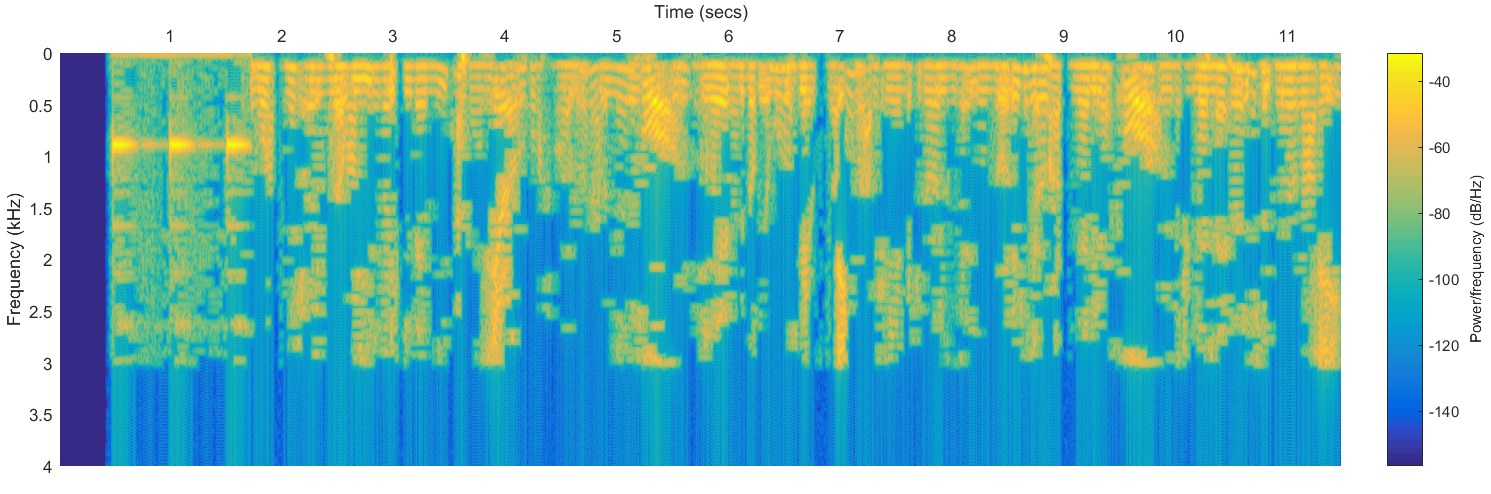}
  \caption{Time-Frequency Analysis.}
\end{figure}

\begin{figure}\label{04648 }
  \centering
  \includegraphics[width=0.7\textwidth]{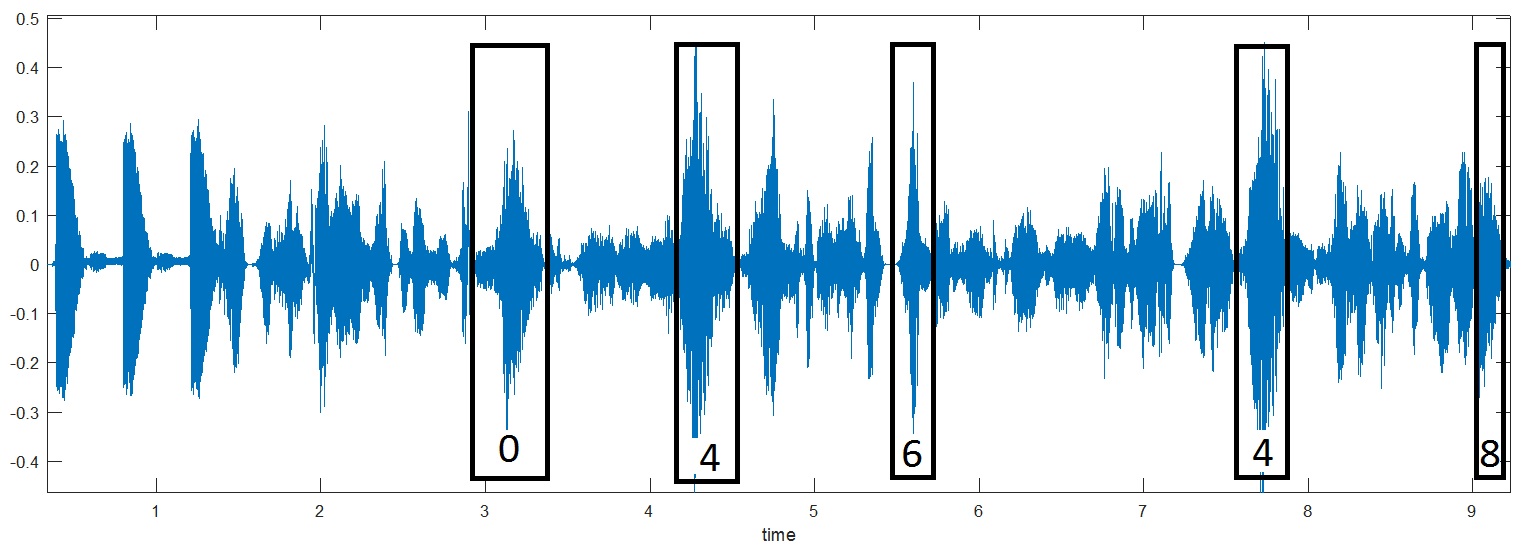}
  \caption{Close look at an audio sample file.} 
\end{figure}

\begin{figure}
  \centering
  \includegraphics[width=0.7\textwidth]{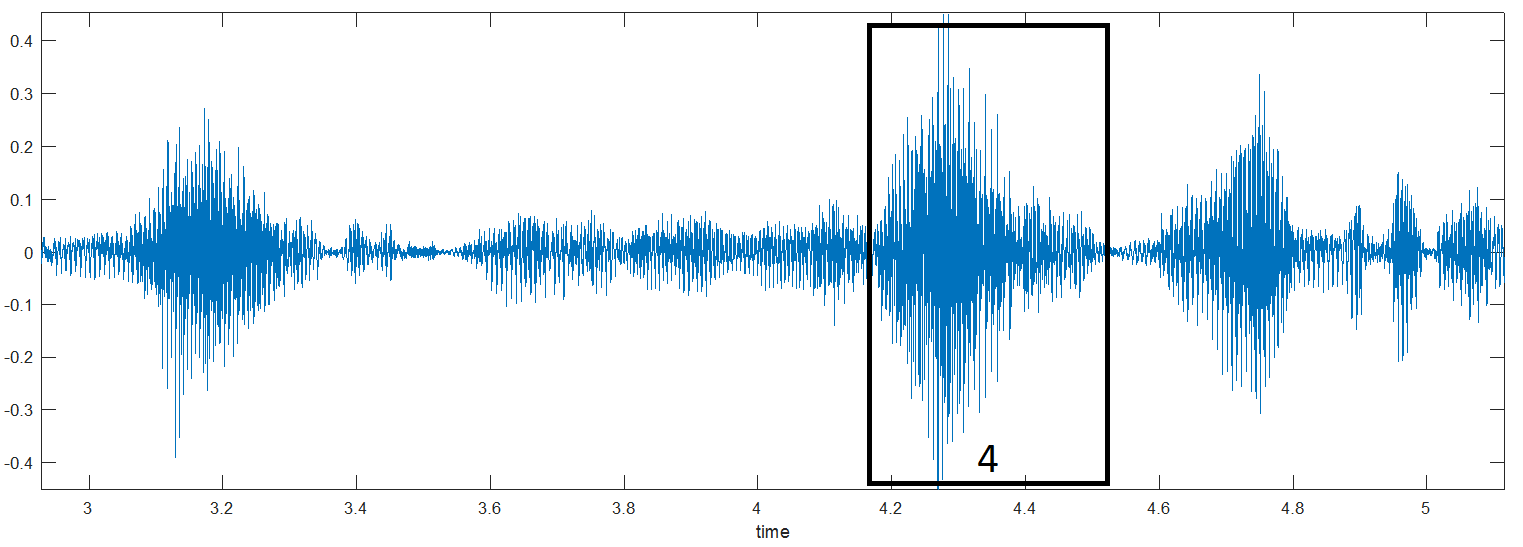}
  \caption{Caption of a particular digit.} 
\end{figure}

\subsection{Algorithms and Techniques}

The Matlab functions for extracting features were published and distributed worldwide at [\hyperref[sec:r15]{15}] and as part of the Voicebox package. We utilize SVM algorithms to carry out automatic digit recognition. We will explain our performance in detail.

The benefits of using SVM library on Python Scikit-learn is that they allow the immediate training of high amount of linear classifiers. Another advantage of SVMs is that they can be kernelled to solve non-linear classification tasks conveniently. The main idea of kernel method to handle linearly inseparable data is to build non-linear combinations of the basic features to project them on to a higher dimensional space via mapping a function $\varphi$ that will be linearly separable. To solve SVM problems, we need to transform our training set on to a higher dimensional feature space using the function $\varphi$ and train a linear SVM model to classify the data in our new feature space. Therefore, we may use that same $\varphi$ to transform the unknown data to classify it by linear SVM model. 

For our problem, we divide the files of the training set into 11 separate classes as 0, 1, 2, ... , 9 and the noise class. We mentioned in Section 3 how many data points we have from each class. Then we extracted the features of these audio signals by using RASTA-PLP. As a feature of RASTA-PLP we obtained $13\times 42$ dimensional feature space. In this number of dimension, 13 is the number of RASTA-PLP  coefficients and 42 is the number of sliding windows that is, every digit have been scanned by 42 sliding windows, and then 13 coefficients have been calculated from each sliding window. Once we calculated the RASTA-PLP coefficients of the digits where they belong to the generated 11 classes, we apply PCA to reduce dimensionality, and then we made a classification using multiclass SVMs.

Henceforth, an existing test audio file has been fragmentized into sliding windows as we did for the training files before. Then, it has been extracted 546 RASTA-PLP features for each sliding window and we used PCA to reduce the dimensionality and SVM to classify these reduced number of features. If the output of the classification results a digit from 0 to 9, then we assigned it as an element of the digit class. Otherwise, if the output results a noise or an awkward silence, then we assigned it as an element of the so-called $11^{th}$ class. 

\maketitle{$SVMs$}

SVMs are supervised machine learning algorithm that can be used for both classification and regression problems. For a determined multiple classes of labeled data, SVMs perform as a distinctive classifier via defining an optimal hyperplane(s) that seperate(s) all classes.

For a given training dataset of n points of the form, assume that the dataset can be linearly separable. Then, 

\begin{equation}
({\vec {x}}_{1},y_{1}),\,\ldots ,\,({\vec {x}}_{n},y_{n})
\end{equation}
where $y_{i}\in\{-1,1\}$ indicates the class if the point ${\vec {x}}_{i}$ belongs to it. The objective is finding the maximum-margin hyperplane that divides the class of points  ${\vec {x}}_{i}$  where $y_{i}=1$ from the class of points where $y_{i}=-1$. The equation of any hyperplane can be written as 
\begin{equation}
{\vec {w}}\cdot {\vec {x}}-b=0,
\end{equation}
where $\vec {w}$ is the normal vector to the hyperplane. 

Since we are dealing with a non-linear classification, the clever way to obtain the classifier is using a kernel trick which applied by Aizerman et al. \cite{aizerman1964theoretical} originally. We will use the Gaussian radial basis function where 
\begin{equation}
{\displaystyle \Phi({\vec {x_{i}}},{\vec {x_{j}}})=\exp(- \frac{\|{\vec {x_{i}}}-{\vec {x_{j}}}\|^{2}}{{2\sigma ^{2}}})} = \exp(-\gamma \|{\vec {x_{i}}}-{\vec {x_{j}}}\|^{2} )
\end{equation}
for $\gamma>0.$

The optimization problem for the hard margin, which is the distance from the hyperplane for the SVM’s decision boundary to the nearest training data point, SVM can be shown by
\begin{equation}
\min_{{\vec {w}},b}\frac{1}{2}||{\vec {w}}||_{2}^{2}
\end{equation}
subject to
\begin{equation}
y_{i}({\vec {w}}^{T} \cdot x_{i}+b)\geq1
\end{equation}
This problem essentially explains us to maximize the margin, while assuring to classify each sample safely and correctly. In practice, it is very hard to apply it, and since our data is not linearly separable, the optimization problem won't give us a solution for parameters such as $w$ and $b$. To deal with this problem, we introduce slack variables such that they allow us to classify each training sample safely and correctly. 

Usually, slack variables are represented by $\xi$ and indicates how much we need to move our sample point in order to classify it safely and correctly. Then we penalize the objective function once we use slack variables, or else we assign the slack variables as high as we can and solve the optimization problem and obtain the following solution:
\begin{equation}
\min_{{\vec {w}},b}\frac{1}{2}||{\vec {w}}||_{2}^{2}+C\cdot\sum_{i=1}^{N}\xi_{i} 
\end{equation}
subject to 
\begin{equation}
y_{i}({\vec {w}}^{T}\cdot x_{i} + b) \geq 1 - \xi_{i}
\end{equation}
where $C$ is the hyperparameter which tests how much we could penalize the usage of slack variables. The process is basically a trade-off between penalizing slacks and attaining a large margin for the SVM.  That is, if we set
\begin{equation}
C \rightarrow 0,
\end{equation}
then we won't penalize the slack variables, and conversely, once we increase the value of $C$, we further penalize the slack variables. In our model, $C$ is the penalty parameter and we took default values for the rest of the SVM parameters. 

We used Gaussian RBF in this problem that is very powerful method since we can transform the inner product with the kernel function $\Phi({\vec {x_{i}}},{\vec {x_{j}}})$.  We may assuming that $x_{i}$ is from the training set and $x_{j}$ is from the testing set. The value of the kernel depends on the $\ell_{2}$-distance between vectors and measures the similarity between two data points. Since $\ell_{2}$ is an infinite dimensional space the kernel $\Phi$ corresponds to a feature space transformation onto an infinite dimensional space. The hyper parameter $\gamma$ manages the trade-off between the error because of bias and the variance in our model. The larger $\gamma$-value implies the smaller $\Phi$-value, even though our input vectors have higher similarity. If the $\gamma$-value is unnecessary large, then we have a big risk of overfitting and prone to high variance. Conversely, a smaller $\gamma$-value leads that the $x_{i}$ vectors (support vectors) have stronger effect on the $x_m$ vectors (classification vectors).

We used SVM with Radial Basis Function and separate the dataset into multi-class SVMs by one to one diagram. We did not take a constant penalty parameter $C_i$ where $i\in\{1,2,\cdots,n\}$, we explored the optimum penalty parameter by applying cross validation. Moreover, we used Naive Bayes algorithm as a baseline model to compare the performances wit our proposed model.   

The proposed model has two independent variables as the coefficient of penalty of SVM and the variance parameter of the PCA. These two values were tuned in the train set by cross-validation. We obtained as optimum values of Penalty Parameter = 50, PCA Var = 0.9 and we decided to use tese particular coefficients when we implement our proposed model.  

\subsection{Benchmark Model}

Without using PCA, we consider naive bayes method and SVM as benchmark models to compare our proposed model. Naive Bayes methods are a set of supervised learning algorithms based on applying Bayes’ theorem with the naive assumption of independence between every pair of features. These classifiers have worked well in many problems, such as text classification and spam filtering. They do not require a large amount of training data to estimate the involved parameters. Naive bayes method can be faster than SVMs when you compare them.  On the other hand, although they are know as a smooth classifiers we can not say they are good estimator. Thus, we applied PCA to our SVM algorithm. 

\section{The Methodology of our Model}

\subsection{Data Preprocessing}

In the proposed algorithm, preprocessing is used to determine the phonemes that are likely to be digitized. 
Therefore, there is no need for a preprocessing step in the train set, since the starting points for each digit set plus the non-digit set of $11^{th}$ are manually set.
When extracting the features of the test set, it is necessary to automatically determine the regions that are likely to be digitized.
Because now nobody has the possibility to run these zones manually.
We implemented a pre-processing algorithm for this.
During this preprocessing phase, the test audio signal is read, 0 is averaged, the signal's energy is calculated, and then the 100-point mean running average of the energy signal is calculated.
The smoothed energy sequence is then run to determine the start of the potential digit regions with some hard thresholding operation.
For this, the smoothed energy sequence is called cluster 1 for large parts from 0 to some degree (larger than 0.00001), and cluster 2 for parts smaller than 0.001.
Then all the start-end point pairs are calculated for the potential digit blanks that start with the element of cluster 1 and end with the element of cluster 2.

In Figure 4, we give the graph of the smoothed signal calculated for the sample audio file and showing the potential starting points.

\begin{figure}
  \centering
    \includegraphics[width=0.87\textwidth]{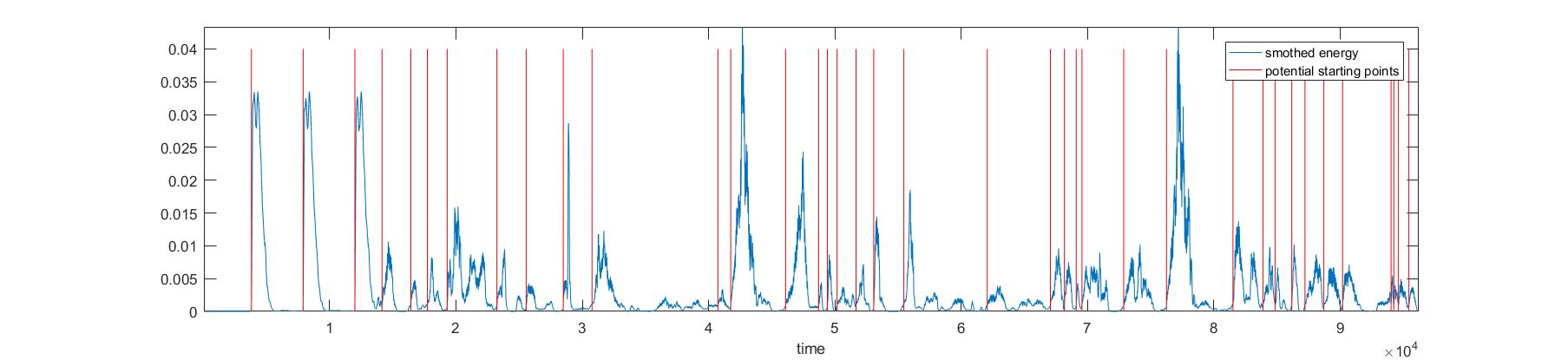}
  \caption{Smoothed signal calculated for the sample audio file and the potential starting points. }
\end{figure}

\subsection{Implementation \& Refinement }

We tested the accuracy of audio CAPTCHAs used by popular machine learning techniques algorithmically planned to break them.

Two different measurements were used for the accuracy of the classification. First, we considered the digits independently to obtain accurate digit prediction. Secondly, we measured the prediction accuracy of the digits where they are usually varying between 4 and 6 for the test files. We operated DTW algorithm to perform the difference between the ground truth and predicted digits.

For feature extraction we use RASTA-PLP  speech analysis [\hyperref[sec:r17]{17}] by applying the following steps: 
\begin{itemize}
\item Calculate the crucial-band spectrum (as in the PLP) and take its log. 
\item Approximate the temporal derivative of log crucial-band spectrum using 4 consecutive spectral values. 
\item Apply SVM as nonlinear classifier for threshold filtering.
\item Integrate log crucial-band temporal derivative.
\item According to ordinary PLP, add equal noise and multiply by 0.33 to create the power law of hearing
\item Operate exponential function of this log spectrum to produce audio spectrum. 
\end{itemize}

To reduce the dimensionality, we used PCA since redundancy were growing as the dimension of the data increased. PCA is one of the most common method to utilize dimensions by transforming them to a new space. We applied k-fold cross validation (in this case $k=4$) to decide the dimension for the usage of the classifier and found the optimum penalty parameter $C$ for the SVM algorithm. The variance parameter used by the PCA algorithm and the k-fold cross validation of the train set were used to determine the best value of the penalty parameter that the SVM algorithm would need.

When subtracting the features of the train set, a signal of 0.4 seconds is taken from each digit, plus the manually set start point for the non-digits of the eleventh set. It is assumed here that each digit is read and finished in 0.4 seconds, which is quite sufficient.
This signal is divided into 42 sliding sub-windows that overlap 50\% and 13 random-plp coefficients of each sub-window are calculated.
Then these coefficients are combined for 42 sub windows to obtain $42 \times13 = 546$ feature vectors.
Now an audio domain, labeled as one of 11 classes, is now expressed in 546 numbers, so its features have been extracted.

Before we extract the features of a test element, the data pre-processing determines the potential start point of the audio block.
Then, from each starting point, another 0.4 second segment is taken and divided into 42 sliding windows overlapping 50\% in the same way as train data.
13 random-plp coefficients of each sub-window are calculated. Then, these coefficients are combined for 42 sub-windows to obtain $42 \times 13 = 546$ feature vectors.
We now have a 546-length feature vector of this potential phoneme.
This 546-length vector is classed according to the method to be applied and is determined to be one of $11^{th}$ class according to the output of the method.
Refinement executed by conducting cross validation so as to determine the most optimum values of PCA variance and C-value.

\section{Numerical Results and Statistics}

\subsection{Model Evaluation \& Validation}

The 4-fold cross validation was applied to the train set to determine the best value of the penalty parameter that the SVM algorithm would need with the variance variable used by the PCA algorithm. According to our algorithm, the entire train set is divided into 4 parts randomly, one part is tested and the remaining 3 parts are accepted as train. The system is trained with the given parameters, then the digit performance for the set determined. In this study, we used 11 different penalty parameters of 1, 10, 20, 30, 40, 50, 60, 70, 80, 90, 100, and 7 different penalty values of 0.25, 0.5, 0.6, 0.7, 0.8, 0.9, 0.99 where for each penalty value we used the value of the variance representation. As a result, 4-fold validation was performed separately for 77 different parameters and the validation success was calculated for each case.

The Table~\ref{tab:table}  shows that the calculated validation accuracies for the varying penalty parameter and PCA variance. As you can see from there, the highest validation success was calculated for Penalty Parameter = 50, PCA-Var = 0.9. 

\begin{table}
 \caption{PCA variance and penalty parameters}
  \centering
  \begin{tabular}{l c  c  c   c  c  c  c  c  c  c  c  c }
    \toprule
    \multicolumn{12}{c}{Penalty Parameter}                   \\
    \cmidrule(r){2-12}
    PCAVar     & 1  & 10 & 20 & 30 & 40 & 50 & 60 & 70 & 80 & 90     & 100 \\
    \midrule
    0.25   & 0.4591      & 0.4641   &0.4655  &0.4656  &0.4658  &0.4669  &0.4672 &0.4672 &0.4671 &0.4674 &0.4674  \\
    0.5     & 0.7808      & 0.8281   &0.8338  &0.8331  &0.8339  &0.8350  &0.8344 &0.8347 &0.8339 &0.8337 &0.8335  \\
    0.6     & 0.8487      & 0.8924   &0.8937  &0.8919  &0.8910  &0.8882  &0.8877 &0.8889 &0.8881 &0.8881 &0.8872 \\
    0.7     & 0.8914      & 0.9240   &0.9205  &0.9189  &0.9183  &0.9166  &0.9158 &0.9156 &0.9155 &0.9150 &0.9155 \\
    0.8     & 0.9058      & 0.9372   &0.9364  &0.9352  &0.9331  &0.9323  &0.9314 &0.9312 &0.9315 &0.9313 &0.9312 \\
    0.9     & 0.8894      & 0.9418   &0.9428  &0.9421  &0.9406  & \textbf{0.9478}  &0.9374 &0.9371 &0.9362 &0.9362 &0.9353 \\
    0.99   & 0.8627      & 0.9360   &0.9311  &0.9316  &0.9323  &0.9318  &0.9367 &0.9366 &0.9368 &0.9373 &0.9368 \\
    \bottomrule
  \end{tabular}
  \label{tab:table}
\end{table}

\begin{table}
 \caption{Train Set Accuracy For Each Class}
  \centering
  \begin{tabular}{c  c  c c  }
          
    \midrule
    Class   & Naive Bayes      & Default SVM   &Proposed SVM   \\
    \toprule
    0     & 0.8787      & 0.8867  &1.0000    \\
    1     & 0.8712      & 0.9738  &1.0000    \\
    2     & 0.8108      & 0.6108  &0.9945    \\
    3     & 0.8735      & 0.9005  &1.0000    \\
    4     & 0.9275      & 0.9565  &1.0000    \\
    5     & 0.9188      & 0.9524  &1.0000    \\
    6     & 0.5090      & 0.2143  &0.9864    \\
    7     & 0.8994      & 0.9143  &1.0000    \\
    8     & 0.6452      & 0.5635  &0.9497    \\
    9     & 0.9193      & 0.9418  &1.0000    \\
 Other  & 0.7234      & 0.9985  &0.9993    \\
    \bottomrule
  \end{tabular}
  \label{tab:table2}
\end{table}

\begin{table}
 \caption{Test Set Accuracy For Digit Based and Entire Sequence}
  \centering
  \begin{tabular}{c  c  c c  }
          
    \midrule
    Type   & Naive Bayes      & Default SVM   &Proposed SVM   \\
    \toprule
    Digit Accuracy     & 0.4213      & 0.9220  & \textbf{0.9809}     \\
    Captcha Accuracy     & 0.0000      & 0.4400  & \textbf{0.8900}     \\    
    \bottomrule
  \end{tabular}
  \label{tab:table3}
\end{table}

\section{Conclusion}

\subsection{Reflection}

The Naïve Bayes method correctly identifies about 42\% of the test digits. This method also failed because every class element in the train set is not balanced because the train set has a large number of noise class ($11^{th}$ class) elements, while the elements from 0 to 9 are somewhat less. None of the 100 audio files in the test set were fully recognized by Naïve Bayes method. Since the number of elements is not balanced, even though the classes from 0 to 9 a are partially recognized even in the train set, the noise class has a relatively low success rate of 71\%. This is why all test audio files are not recognized without error, yet even if a section is wrongly assigned to the noise class, it means that the test element is misclassified

On the other hand, when the default number of SVM elements is unstable, it has succeeded in achieving 92\% independent digits, but only 44\% of the test elements have been achieved. With the parameter we have optimized, the recognition of the complex chaptche sound file has reached 89\%. Independent digit success has reached a very good value of 98\%.

We applied 3 different algorithms naive bayes (no PCA), default SVM (no PCA) and SVM (with PCA) to optimize the parameters as promised. The naive bayes method failed because each class element in the train set was not balanced. Because there are a large number of noise classes in the train cluster, the elements from 0 to 9 are somewhat less. None of the 100 audio files in the test set were fully recognized by Naïve Bayes. 

Once we compare the results of our algorithm for each case we found the following results in Table~\ref{tab:table2} and Table~\ref{tab:table3} for train and test set respectively.

\subsection{Improvement}

We have accomplished in decoding a certain type of audio CAPTCHAs by machine learning techniques. It is obvious that our results can be improved by rearranging appropriate segments and algorithms. We have proved that our approach is is working out and might be used with various different audio CAPTCHAs that involve finite conversations and noise.

\section*{References}

\medskip

\small

[1] \label{sec:r1} L.~Ahn, M.~Blum, N.~Hopper, and J.~Langford, ``Using hard ai problems for
  security,'' in {\em Proceedings of the 22nd international conference on
  Theory and applications of cryptographic techniques}, pp.~294--311, 2007.

[2] \label{sec:r2}H.~S. Baird, A.~L. Coates, and R.~J. Fateman, ``Pessimalprint: a reverse turing
  test,'' {\em International Journal on Document Analysis and Recognition},
  vol.~5, no.~2-3, pp.~158--163, 2003.

[3] \label{sec:r3}M.~Chew and H.~S. Baird, ``Baffletext: A human interactive proof,'' in {\em
  Document Recognition and Retrieval X}, vol.~5010, pp.~305--317, International
  Society for Optics and Photonics, 2003.

[4] \label{sec:r4} L.~Von~Ahn, M.~Blum, and J.~Langford, ``Telling humans and computers apart
  automatically,'' {\em Communications of the ACM}, vol.~47, no.~2, pp.~56--60,
  2004.

[5] \label{sec:r5}J.~Elson, J.~J. Douceur, J.~Howell, and J.~Saul, ``Asirra: A captcha that
  exploits interest-aligned manual image categorization,'' in {\em Proceedings
  of the 14th ACM Conference on Computer and Communications Security}, CCS '07,
  (New York, NY, USA), pp.~366--374, ACM, 2007.

[6] \label{sec:r6}G.~Sauer, H.~Hochheiser, J.~Feng, and J.~Lazar, ``Towards a universally usable
  captcha,'' in {\em Proceedings of the 4th Symposium on Usable Privacy and
  Security}, vol.~6, p.~1, 2008.

[7] \label{sec:r7}J.~Tam, J.~Simsa, S.~Hyde, and L.~V. Ahn, ``Breaking audio captchas,'' in {\em
  Advances in Neural Information Processing Systems}, pp.~1625--1632, 2009.

[8] \label{sec:r8}H.~Gao, H.~Liu, D.~Yao, X.~Liu, and U.~Aickelin, ``An audio captcha to
  distinguish humans from computers,'' in {\em Electronic Commerce and Security
  (ISECS), 2010 Third International Symposium on}, pp.~265--269, IEEE, 2010.

[9] \label{sec:r9} G.~Mori and J.~Malik, ``Recognizing objects in adversarial clutter: Breaking a
  visual captcha,'' in {\em Computer Vision and Pattern Recognition, 2003.
  Proceedings. 2003 IEEE Computer Society Conference on}, vol.~1, pp.~I--I,
  IEEE, 2003.

[10] \label{sec:r10} K.~Chellapilla and P.~Y. Simard, ``Using machine learning to break visual human
  interaction proofs (hips),'' in {\em Advances in neural information
  processing systems}, pp.~265--272, 2005.

[11] \label{sec:r11}H.~Sakoe and S.~Chiba, ``Dynamic programming algorithm optimization for spoken
  word recognition,'' {\em IEEE transactions on acoustics, speech, and signal
  processing}, vol.~26, no.~1, pp.~43--49, 1978.

[12] \label{sec:r12}H.~Hermansky, ``Perceptual linear predictive (plp) analysis of speech,'' {\em
  the Journal of the Acoustical Society of America}, vol.~87, no.~4,
  pp.~1738--1752, 1990.

[13] \label{sec:r13}H.~Hermansky, N.~Morgan, A.~Bayya, and P.~Kohn, ``Rasta-plp speech analysis
  technique,'' in {\em Acoustics, Speech, and Signal Processing, 1992.
  ICASSP-92., 1992 IEEE International Conference on}, vol.~1, pp.~121--124,
  IEEE, 1992.

[14] \label{sec:r14}J.~Tam, J.~Simsa, D.~Huggins-Daines, L.~Von~Ahn, and M.~Blum, ``Improving audio captchas,'' in {\em Symposium On Usable Privacy and Security (SOUPS)}, 2008.

[15] \label{sec:r15} D.~Ellis, ``Plp and rasta (and mfcc, and inversion) in matlab using melfcc. m
  and invmelfcc. m,'' 2006.

[16] \label{sec:r16}M.~A. Aizerman, ``Theoretical foundations of the potential function method in
  pattern recognition learning,'' {\em Automation and remote control}, vol.~25,
  pp.~821--837, 1964.

[17] H.~Morgan, N.~Bayya, A.~Kohn, and P.~Hermansky, ``Rasta-plp speech analysis,''  {\em ICSI Technical Report TR-91-969}, 1991. \label{sec:r17}

\end{document}